\documentclass{article}

\usepackage{arxiv}

\usepackage[utf8]{inputenc} 
\usepackage[T1]{fontenc}    
\usepackage{hyperref}       
\usepackage{url}            
\usepackage{booktabs}       
\usepackage{amsfonts}       
\usepackage{nicefrac}       
\usepackage{microtype}      
\usepackage{lipsum}
\usepackage{graphicx}
\graphicspath{ {./images/} }

\usepackage{caption}
\usepackage{subcaption}

\usepackage{floatrow}
\newfloatcommand{capbtabbox}{table}[][\FBwidth]

\usepackage{blindtext}

\title{Domain-adapted large language models for classifying nuclear medicine reports}

\author{
 Zachary Huemann \\
  University of Wisconsin-Madison\\
  Madison, WI, USA\\
  \texttt{zhuemann@wisc.edu} \\
   \And
 Changhee Lee \\
  University of Wisconsin-Madison\\
  Madison, WI, USA\\
  \texttt{changhee.lee@wisc.edu} \\
  \And
 Junjie Hu \\
  University of Wisconsin-Madison\\
  Madison, WI, USA\\
  \texttt{junjie.hu@wisc.edu} \\
    \And
 Steve Y. Cho \\
  University of Wisconsin-Madison\\
  Madison, WI, USA\\
  \texttt{cho85@wisc.edu} \\
    \And
 Tyler Bradshaw \\
  University of Wisconsin-Madison\\
  Madison, WI, USA\\
  \texttt{tbradshaw@wisc.edu} \\
}

\begin{document}
\maketitle
\begin{abstract}
With the growing use of transformer-based language models in medicine, it is unclear how well these models generalize to nuclear medicine which has domain-specific vocabulary and unique reporting styles. In this study, we evaluated the value of domain adaptation in nuclear medicine by adapting language models for the purpose of 5-point Deauville score prediction based on clinical 18F-fluorodeoxyglucose (FDG) PET/CT reports. We retrospectively retrieved 4542 text reports and 1664 images for FDG PET/CT lymphoma exams from 2008-2018 in our clinical imaging database. Deauville scores were removed from the reports and then the remaining text in the reports was used as the model input. Multiple general-purpose
transformer language models were used to classify the reports into Deauville scores 1-5. We then adapted the models to the nuclear medicine domain using masked language modeling and assessed its impact on classification performance. The language models were compared against vision models, a multimodal vision language model, and a nuclear medicine physician with seven-fold Monte Carlo cross validation, reported are the mean and standard deviations. Domain adaption improved all language models. For example, BERT improved from 61.3 ± 2.9\% fiveclass accuracy to 65.7 ± 2.2\% following domain adaptation. The best performing model (domain-adapted RoBERTa) achieved a five-class accuracy of 77.4 ± 3.4\%, which was better than the physician’s performance (66\%), the best vision model’s performance (48.1 ± 3.5\%), and was similar to the multimodal model’s performance (77.2 ± 3.2\%). Domain adaptation improved the performance of large language models in interpreting nuclear medicine text reports. 
\end{abstract}


\section{Introduction}
A\footnote{This work has been submitted to Radiology: Artificial Intelligence for possible publication. Copyright may be transferred without notice, after which this version may no longer be accessible.} radiology report is the culmination of a radiological exam. It represents an official interpretation of the patient’s images and aims to provide actionable information for future care. Given the vast number of information-rich radiology reports stored in clinical imaging databases, it is no surprise that artificial intelligence (AI)-based language models have begun to find an increasing number of applications in radiology. Language models can extract training labels from reports \cite{smit-etal-2020-combining}\cite{tejani_performance_2022}\cite{Fink_oncology}, summarize reports \cite{yan_radbert_2022}, generate reports from images \cite{yan_a_contrastive}, and more \cite{sorin_deep_2020}. Despite their impressive performance in various domains, language models trained on large generic text corpora can be suboptimal for applications in radiology and nuclear medicine. Large pre-trained language models, such as Bidirectional Encoder Representations from Transformers (BERT) \cite{devlin_bert_nodate}, are typically developed using selfsupervised training, which involves predicting occluded words or adjacent sentences within the training corpus. Adapting general-purpose language models to a particular domain by subjecting it to additional self-supervised training using domain-specific text has been shown to boost performance for certain tasks \cite{lee_biobert}. For example, bioBERT was adapted from BERT through additional pre-training on biomedical corpora from PubMed \cite{lee_biobert}. Domain-adapted bioBERT outperformed BERT when applied to downstream biomedical tasks. Other studies have demonstrated the benefit of domain adaption \cite{chaudhari_application_2022} \cite{khare_MMBERT}.

\begin{figure*}[!h]
\centerline{\includegraphics[width=0.8\textwidth]{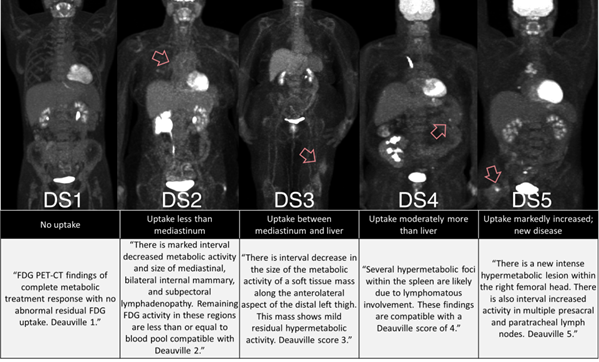}}
\caption{The prediction task was the five-point scale of the Deauville criteria for clinical reporting of PET-based treatment response of lymphoma. Example text descriptions from radiology reports and maximum intensity projection images are shown for each of the 5 Deauville (DS) categories.}
\label{fig1}
\end{figure*}

Adaptation of language models to the nuclear medicine domain, however, has been understudied. Nuclear medicine reports contain unique terms that are not used more broadly in radiology. For example, “hypermetabolic”, “focal uptake”, and “SUVmax” are esoteric terms but are critically important to the interpretation of a nuclear medicine report. It is unclear if generic or biomedical-adapted language models can capture the underlying meaning of nuclear medicine vocabulary. Furthermore, it is uncertain how much can be gained by domain adaption. In this feasibility study we evaluated the performance of large language models in interpreting and classifying PET/CT text reports. The prediction task that we studied was the five-point scale of the Deauville criteria. Deauville scores (DS) are routinely used for clinical reporting of PET-based diagnosis and response assessment for lymphoma Fig.~\ref{fig1}. We explored DS prediction as it is representative of other similar interpretive tasks, such as diagnosis extraction, and requires the language model to learn global interpretations of the free-text reports. We compared the classification performance of language models to a human expert, vision models, and a multimodal model. We hypothesised that domain adaptation to nuclear medicine will improve performance on the downstream task of predicting DS. 

\section{Methods}

\subsection{Data Collection}

Using an institutional review board-approved retrospective protocol with waiver of informed consent, we queried the universities picture archiving and communication system (PACS) for clinical 18F-fluorodeoxyglucose (FDG) PET/CT exams that contained the term “lymphoma” in the exam indication or impression. A total of 4,542 exams dictated by at least 44 different physicians, 12 radiology faculty and 32 residents, were identified between the years 2008-2018 and their images were anonymized using ClinicalTrialProcessor and downloaded together with their radiology reports.

\subsection{Deauville Score Extraction}

PET/CT exams containing a physician-assigned DS in the corresponding report were identified by searching report text for the string “Deauville” and its common misspellings. N-gram analysis was performed to identify all the ways by which DS was reported by physicians (e.g., “Deauville score of 1”, “1 on the Deauville scale”, etc.). Each exam was then assigned a classification label 1-5 according to the DS. If reports contained two or more DSs, such as for lesion-specific reporting, the highest-valued DS was used as the exam-level label. The DS were then redacted from the report and the remaining text was used as input to the language models. 
\begin{table} 
\centering
\caption{Frequency of each Deauville score in the dataset. }
\begin{tabular}{lll}
\hline
Deauville Score & Definition & Frequency   \\
\hline
1               &    No uptake above the background        & 313         \\
2               &  Uptake less than mediastinum          & 355         \\
3               &  Uptake greater than mediastinum but less than or equal to liver          & 155         \\
4               &   Uptake moderately increased compared to the liver at any site         & 221         \\
5               &   Uptake markedly increased compared to the liver at any site         & 620         \\
\hline
total           & --         & 1664\\
\bottomrule
\end{tabular}
\label{table1}
\end{table}

\begin{figure*}[!h]
\centerline{\includegraphics[width=0.3\textwidth]{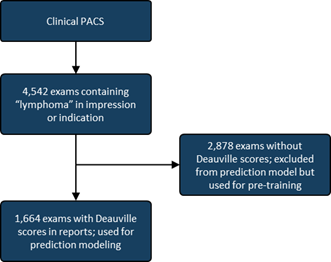}}
\caption{Flow chart of patient inclusion/exclusion.}
\label{fig4}
\end{figure*} 

\subsection{Language Models}

\begin{figure*}[!h]
\centerline{\includegraphics[width=0.9\textwidth]{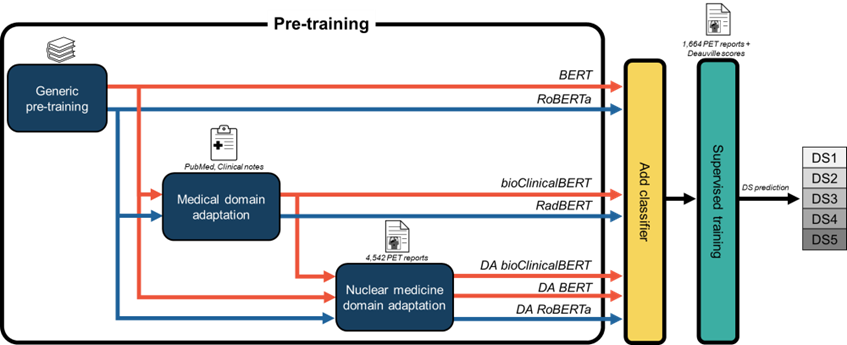}}
\caption{The pre-training and fine-tuning of the various language models are shown. Transformer models trained on generic text (BERT and RoBERTa) were compared to models that had additional pre-training on medical corpora (bioClinicalBERT) and to models specifically domain-adapted (DA) to nuclear medicine (DA bioClinicalBERT, DA BERT, and DA RoBERTa). DA consisted of further pre-training with masked language modeling using text from PET reports. Each model was appended with a classifier and underwent supervised training to predict the 5-class Deauville scores (DS).}
\label{fig2}
\end{figure*}

Four transformer-based language models were evaluated Fig.~\ref{fig2}. This included BERT, a 110M parameter transformer model with 12 transformer blocks and 12 self-attention heads pretrained on BooksCorpus and English Wikipedia using two pretraining tasks: masked language modeling (MLM) and next sentence prediction (NSP) \cite{devlin_bert_nodate}. We also evaluated BioClinicalBERT \cite{alzentzer_BIOclinicalBERT}, which was initialized from BioBERT \cite{lee_biobert} and then further pretrained using clinical notes (880M words)\cite{johnson_mimic-iii_2016}. RadBERT is a 123M parameter model initialized from RoBERTa-Base and pretrained on 4M radiology reports \cite{yan_radbert_2022}. Finally, we evaluated RoBERTa \cite{liu_roberta}, a 355M parameter model with 24 transformer blocks and 16 self-attention heads pretrained using dynamic MLM. All pre-trained language models were evaluated for their ability to classify radiology reports into DS categories 1-5. The redacted PET/CT text reports were preprocessed with punctuation removal, date stripping, and numerical rounding. We performed synonym replacement to create homogeneity between different physicians’ vocabularies using a custom list of synonyms (e.g., “SUV” = “SUVmax” = “standardized uptake value”), but later found this had no measurable impact on model performance. Due to the language models’ 512-token limit on the input text, we prioritized the impression section as input and then included as much of the findings section as was allowed. This is because impression sections generally summarize the findings sections. Input text was converted into word embeddings using subword tokenization and fed to the pre-trained language models. The output of each language model was used as input to a 3-layer classifier with two fully connected layers (768 nodes for BERT and 1024 for RoBERTa) and a final softmax layer. The classifiers were trained using cross-entropy loss with an ADAM optimizer. All models were trained and evaluated using seven iterations of random sampling cross validation with splits of 80\% training, 10\% testing, and 10\% validation (used for early stopping). All language models were imported from the Hugging Face library \cite{wolf_transformers_2020} and implemented in PyTorch.

\subsection{Domain Adaptation}

We evaluated the impact of domain adaptation on the classification performance of the language models. We performed additional MLM pre-training using the 4,542 PET/CT reports as the pre-training corpus (2M words). MLM pretraining consisted of randomly masking 15\% of the tokens and predicting the masked tokens. We pretrained the model for 3 epochs using a learning rate of 1e-6, which prevented overfitting. Classification models incorporating the domain-adapted language models were then fine-tuned according to the same training procedure as described above. 

\subsection{Comparator Methods}

We trained vision models that used PET images as input for comparison. We trained a vision transformer (ViT) \cite{dosovitskiy_vit} and an EfficientNet-B7 convolutional neural network \cite{tan_efficient_net} to predict DS using corresponding coronal maximum intensity projection (MIP) PET images as input. The ViT model consisted of 12 transformer layers with 12 attention heads. We pretrained our models using ImageNet-21k \cite{ridnik_imagenet_21k}. Images were cropped to the thighs and resized to 384x384 with pixel normalization. We used standard augmentations: horizontal flipping, vertical flipping, random rotation, and random translation. 

We also compared the performance of the language and vision models to a human expert. A nuclear medicine physician with 4 years of nuclear medicine experience was given the redacted reports and MIP PET images of 50 random cases and was asked to predict the DS that had been originally assigned by the reading physician.

\begin{figure}
\begin{floatrow}
\ffigbox{%
  \includegraphics[width=0.4\textwidth]{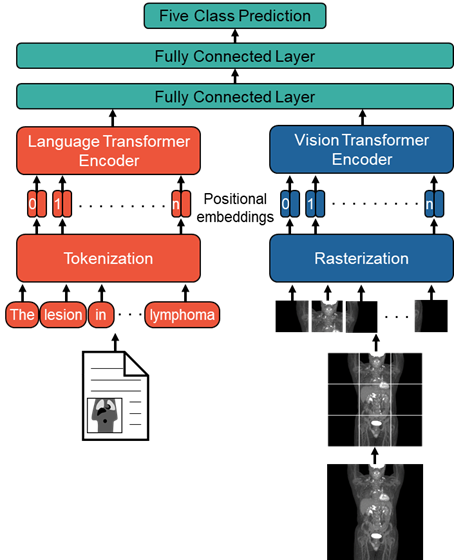}%
}{%
  \caption{The multimodal model consisted of two mode-specific pathways: one based on a RoBERTa language transformer and the other based on a vision transformer (ViT). Embeddings from both pathways were concatenated and passed through a 3-layer classifier. \label{fig3}}%
}
\capbtabbox{%
  \begin{tabular}{lll}
  \hline
\multicolumn{1}{p{2cm}}{\centering Method} & \multicolumn{1}{p{2cm}}{\centering Weighted Cohen's $\kappa$} & \multicolumn{1}{p{2cm}}{\centering 5-Class \\ Accuracy (\%)}   \\
\hline
BERT             &   0.65       & 61.3 $\pm$ 2.9   \\
RoBERTa          &  0.81         & 73.7 $\pm$ 4.0   \\
bioClinicalBERT  &  0.68         &  63.0 $\pm$ 3.6   \\
RadBERT          &  0.81         &   73.0 $\pm$ 1.9  \\
\hline
DA BERT          & 0.70           & 65.7 $\pm$ 2.2       \\
DA RoBERTa       & 0.83       & 77.4 $\pm$ 3.4          \\
DA bioClinicalBERT & 0.74  & 66.4 $\pm$ 3.2   \\
\hline
Vision Transformer  & 0.39 & 46.5 $\pm$ 3.1   \\
EfficientNet        & 0.53 & 47.1 $\pm$ 3.5   \\
\hline
Multimodal          & 0.83 & 77.2 $\pm$ 3.2   \\
\hline
Human Expert        & 0.79 & 66 \\
\bottomrule
\end{tabular}
}{%
  \caption{The performance of different methods on predicting Deauville scores.\label{table2}}
}
\end{floatrow}
\end{figure}

Lastly, we evaluated the performance of a multimodal model that simultaneously operated on paired images and text. We aimed to determine if text reports and images contained complementary or redundant information for the prediction task. Our multimodal model combined ViT and domain-adapted RoBERTa. Embeddings from both models were concatenated and then fed to a 3-layer classifier with two 1024-node layers Fig.~\ref{fig3}.


\section{Results}

Out of the 4,542 PET/CT exams for lymphoma, a total of 1,664 exams contained DSs, with the frequencies of different DS categories shown in Table.~\ref{table1}. Exams not containing DSs were acquired prior to our clinic’s adoption of the Deauville criteria (particularly for baseline scans) or likely had indications other than lymphoma. Patient ages ranged from 18-95 (mean: 53) and 42\% and 58\% were female and male, respectively.

\begin{figure}%
    \centering
    \subfloat[\centering]{{\includegraphics[width=7.5cm]{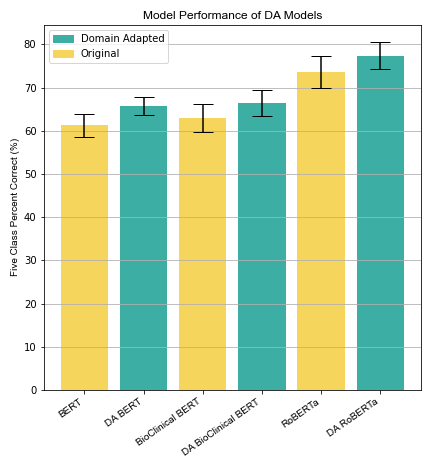} }}%
    \qquad
    \subfloat[\centering]{{\includegraphics[width=7.5cm]{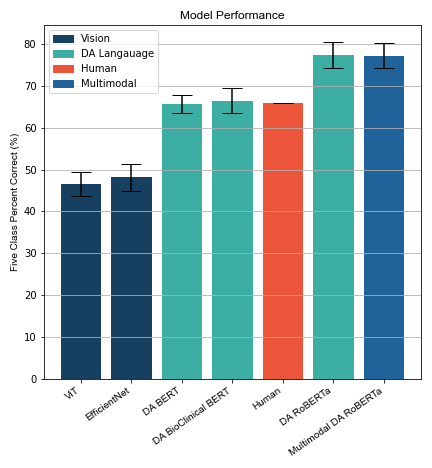} }}%
    \caption{(a): Comparison of domain-adapted (DA) language models and non domain-adapted models. (b): Comparison of domain-adapted models, dision models, a multimodal model, and a human expert in predicting Deauville scores. Shown are five-class accuracy ±SD for each model in comparison to the reported Deauville score.}%
    \label{fig6}%
\end{figure}

The DS classification results are shown for the language models with and without domain adaptation in Fig.~\ref{fig6}. Weighted Cohen’s $\kappa$ are shown in Table.~\ref{table2}. The BERT and RoBERTa models achieved 5-class accuracies of 61.3 ± 2.9\% and 73.7 ± 4.0\%, respectively. BioClinicalBERT performed similarly to BERT: 63.0 ± 3.6\%. Domain adaptation resulted in improved prediction performance for each of the models: BERT improved to 65.7 ± 2.2\%, bioClinicaBERT improved to 66.4 ± 3.2\%, and RoBERTa improved to 77.4 ± 3.4\% (see Fig.~\ref{fig6}). RadBERT had an accuracy of 73.0 ± 1.9\% which is comparable to the larger RoBERTa model. The larger RoBERTa-based models outperformed the BERT models, even outperforming the human expert (66\%) as shown in Fig.~\ref{fig6}.

Vision models performed substantially worse than language models. ViT and EfficientNet achieved accuracies of 46.5 ± 3.1\% and 48.1 ± 3.5\%, respectively. Fig.~\ref{fig6} shows a comparison of vision models and language models. Weighted Cohen’s $\kappa$ are shown in Table.~\ref{table2}.


The multimodal model failed to outperform the domain-adapted RoBERTa language model, which constituted the multimodal model’s language pathway. The multimodal model achieved an accuracy of 77.2 ± 3.2\% whereas the domain-adapted RoBERTa model achieved an accuracy of 77.4 ± 3.4\%, indicating that representations learned from the images did not supplement the representations learned from the text for this particular prediction task. Both models were better than the human expert. 


\section{Discussion}

We found that domain-adapted language models performed better than general-purpose language models at predicting DS from nuclear medicine reports. Performance gains were between 3.4–4.4\% in five-class accuracy. Our DA BERT model used an additional 2M word corpus on top of BERT which yielded a gain of 4.4\% accuracy. BioClincialBERT is a standard BERT model with an additional 880M word corpus of biomedical text which yielded a gain of 1.7\% accuracy. This highlights the advantage of domain specific pretraining in nuclear medicine when compared to general biomedical domain pretraining (e.g., BioClincialBERT). RadBERT was able to perform similarly to the larger RoBERTa model, which contains almost 3 times the number of parameters of RadBERT, by leveraging domain specific pretraining. Given these results and the relative ease with which domain adaptation can be performed, as it does not require data to be labeled and the technical processes have been conveniently streamlined with available Python libraries, we feel that language models intended to operate on nuclear medicine data should undergo domain adaptation. Moreover, language models adapted to the general biomedical domain (e.g., bioBERT) did not convincingly outperform the general-purpose BERT model, and both were inferior to the nuclear medicine domain-adapted models. This suggests that it will be important to include nuclear medicine text in any corpus that is used to adapt language models to the field of radiology.

DS prediction from nuclear medicine reports is arguably a difficult task for language models. While classification of radiology reports according to DS is itself not a clinically useful task, it represents a complex interpretation challenge to language models, and is a problem with conveniently available data labels (i.e., physician-assigned DS) enabling large-scale analysis. We explored this task expecting that it will be representative of other similar interpretive tasks (e.g., diagnosis extraction, report summarization, etc.) that don’t have convenient labels. Unlike some language tasks that allow models to operate more locally on a text input, such as named entity recognition or spelling/grammar checking, DS prediction requires a more global interpretation of the report. Models must be capable of weighing disease-negative sentences (e.g., “prior splenic uptake has resolved”) against diseasepositive sentences (e.g., “new hypermetabolic mediastinal nodes”), as reports often contain a combination of both. Furthermore, the distinction between two consecutive Deauville categories can be minor. Overall, the impressive performance of the language models at this task, with some models exceeding the performance of a human expert, is highly encouraging for their future use in nuclear medicine.

We found that models operating on images alone were less accurate at the DS prediction task than language models. This is unsurprising given that the task was to predict the DS as assigned by the original reading physician who was also the individual who wrote the report, both of which reflect a subjective interpretation of the PET images and not some objective ground truth. We used a multimodal model to determine if the information provided by images could complement the information provided by the language in the report. But we found that the language information dominated the prediction task to the point where the image information provided no additional gains.

There were some notable limitations of the study. First, the reports and images used in this study originated from a single institution and may be lacking in diversity of reporting styles. While the reports in our text corpus were dictated by 44 different physicians, including a mixture of attendings, trainees, nuclear medicine specialists, and dual-trained radiologists, reporting practices at a single institution are likely to be more uniform than across different institutions. Our study was also limited to a single prediction task. Therefore, it is uncertain how well our results would generalize to different institutions and different prediction tasks. Additionally, we only benchmarked against a single human expert, so there is no interobserver agreement rate for comparison of this specific task. Lastly, our PET images were collapsed to 2D MIPs so that we could use existing image classification transformer and convolutional neural networks. MIPs have been found to have prognostic value for AI algorithms \cite{girum_18_2022} but do
have some inherent limitations despite their common use during clinical reads.

In conclusion, we found that large language models are highly capable of interpreting nuclear medicine text reports. Models that were adapted to the nuclear medicine domain via self-supervised learning outperformed general-purpose models, with some models exceeding the prediction performance of a human expert.

\bibliographystyle{unsrt}  
\bibliography{references}  

\begin{thebibliography}{10}

\bibitem{smit-etal-2020-combining}
Akshay Smit, Saahil Jain, Pranav Rajpurkar, Anuj Pareek, Andrew Ng, and Matthew
  Lungren.
\newblock Combining automatic labelers and expert annotations for accurate
  radiology report labeling using {BERT}.
\newblock In {\em Proceedings of the 2020 Conference on Empirical Methods in
  Natural Language Processing (EMNLP)}, pages 1500--1519, Online, November
  2020. Association for Computational Linguistics.

\bibitem{tejani_performance_2022}
Ali~S. Tejani, Yee~S. Ng, Yin Xi, Julia~R. Fielding, Travis~G. Browning, and
  Jesse~C. Rayan.
\newblock Performance of {Multiple} {Pretrained} {BERT} {Models} to {Automate}
  and {Accelerate} {Data} {Annotation} for {Large} {Datasets}.
\newblock {\em Radiology: Artificial Intelligence}, 4(4):e220007, July 2022.

\bibitem{Fink_oncology}
Matthias~A. Fink, Klaus Kades, Arved Bischoff, Martin Moll, Merle Schnell,
  Maike K\"{u}chler, Gregor K\"{o}hler, Jan Sellner, Claus~Peter Heussel,
  Hans-Ulrich Kauczor, Heinz-Peter Schlemmer, Klaus Maier-Hein, Tim~F. Weber,
  and Jens Kleesiek.
\newblock Deep learning–based assessment of oncologic outcomes from natural
  language processing of structured radiology reports.
\newblock {\em Radiology: Artificial Intelligence}, 4(5):e220055, 2022.

\bibitem{yan_radbert_2022}
An~Yan, Julian McAuley, Xing Lu, Jiang Du, Eric~Y. Chang, Amilcare Gentili, and
  Chun-Nan Hsu.
\newblock {RadBERT}: {Adapting} {Transformer}-based {Language} {Models} to
  {Radiology}.
\newblock {\em Radiology: Artificial Intelligence}, 4(4):e210258, July 2022.

\bibitem{yan_a_contrastive}
An~Yan, Zexue He, Xing Lu, Jiang Du, Eric Chang, Amilcare Gentili, Julian
  McAuley, and Chun-Nan Hsu.
\newblock Weakly supervised contrastive learning for chest x-ray report
  generation, 2021.

\bibitem{sorin_deep_2020}
Vera Sorin, Yiftach Barash, Eli Konen, and Eyal Klang.
\newblock Deep {Learning} for {Natural} {Language} {Processing} in
  {Radiology}—{Fundamentals} and a {Systematic} {Review}.
\newblock {\em Journal of the American College of Radiology}, 17(5):639--648,
  May 2020.

\bibitem{devlin_bert_nodate}
Jacob Devlin, Ming-Wei Chang, Kenton Lee, and Kristina Toutanova.
\newblock {BERT}: {Pre}-training of {Deep} {Bidirectional} {Transformers} for
  {Language} {Understanding}.

\bibitem{lee_biobert}
Jinhyuk Lee, Wonjin Yoon, Sungdong Kim, Donghyeon Kim, Sunkyu Kim, Chan~Ho So,
  and Jaewoo Kang.
\newblock {BioBERT: a pre-trained biomedical language representation model for
  biomedical text mining}.
\newblock {\em Bioinformatics}, 36(4):1234--1240, 09 2019.

\bibitem{chaudhari_application_2022}
Gunvant~R. Chaudhari, Tengxiao Liu, Timothy~L. Chen, Gabby~B. Joseph, Maya
  Vella, Yoo~Jin Lee, Thienkhai~H. Vu, Youngho Seo, Andreas~M. Rauschecker,
  Charles~E. McCulloch, and Jae~Ho Sohn.
\newblock Application of a {Domain}-specific {BERT} for {Detection} of {Speech}
  {Recognition} {Errors} in {Radiology} {Reports}.
\newblock {\em Radiology: Artificial Intelligence}, 4(4):e210185, July 2022.

\bibitem{khare_MMBERT}
Yash Khare, Viraj Bagal, Minesh Mathew, Adithi Devi, U~Deva Priyakumar, and
  CV~Jawahar.
\newblock Mmbert: Multimodal bert pretraining for improved medical vqa, 2021.

\bibitem{alzentzer_BIOclinicalBERT}
Emily Alsentzer, John~R. Murphy, Willie Boag, Wei-Hung Weng, Di~Jin, Tristan
  Naumann, and Matthew B.~A. McDermott.
\newblock Publicly available clinical bert embeddings, 2019.

\bibitem{johnson_mimic-iii_2016}
Alistair~E.W. Johnson, Tom~J. Pollard, Lu~Shen, Li-wei~H. Lehman, Mengling
  Feng, Mohammad Ghassemi, Benjamin Moody, Peter Szolovits, Leo Anthony~Celi,
  and Roger~G. Mark.
\newblock {MIMIC}-{III}, a freely accessible critical care database.
\newblock {\em Scientific Data}, 3(1):160035, May 2016.

\bibitem{liu_roberta}
Yinhan Liu, Myle Ott, Naman Goyal, Jingfei Du, Mandar Joshi, Danqi Chen, Omer
  Levy, Mike Lewis, Luke Zettlemoyer, and Veselin Stoyanov.
\newblock Roberta: A robustly optimized bert pretraining approach, 2019.

\bibitem{wolf_transformers_2020}
Thomas Wolf, Lysandre Debut, Victor Sanh, Julien Chaumond, Clement Delangue,
  Anthony Moi, Pierric Cistac, Tim Rault, Remi Louf, Morgan Funtowicz, Joe
  Davison, Sam Shleifer, Patrick von Platen, Clara Ma, Yacine Jernite, Julien
  Plu, Canwen Xu, Teven Le~Scao, Sylvain Gugger, Mariama Drame, Quentin Lhoest,
  and Alexander Rush.
\newblock Transformers: {State}-of-the-{Art} {Natural} {Language} {Processing}.
\newblock In {\em Proceedings of the 2020 {Conference} on {Empirical} {Methods}
  in {Natural} {Language} {Processing}: {System} {Demonstrations}}, pages
  38--45, Online, 2020. Association for Computational Linguistics.

\bibitem{dosovitskiy_vit}
Alexey Dosovitskiy, Lucas Beyer, Alexander Kolesnikov, Dirk Weissenborn,
  Xiaohua Zhai, Thomas Unterthiner, Mostafa Dehghani, Matthias Minderer, Georg
  Heigold, Sylvain Gelly, Jakob Uszkoreit, and Neil Houlsby.
\newblock An image is worth 16x16 words: Transformers for image recognition at
  scale, 2020.

\bibitem{tan_efficient_net}
Mingxing Tan and Quoc~V. Le.
\newblock Efficientnet: Rethinking model scaling for convolutional neural
  networks.
\newblock 2019.

\bibitem{ridnik_imagenet_21k}
Tal Ridnik, Emanuel Ben-Baruch, Asaf Noy, and Lihi Zelnik-Manor.
\newblock Imagenet-21k pretraining for the masses, 2021.

\bibitem{girum_18_2022}
Kibrom~Berihu Girum, Louis Rebaud, Anne-Ségolène Cottereau, Michel Meignan,
  Jérôme Clerc, Laetitia Vercellino, Olivier Casasnovas, Franck Morschhauser,
  Catherine Thieblemont, and Irène Buvat.
\newblock $^{\textrm{18}}$ {F}-{FDG} {PET} maximum intensity projections and
  artificial intelligence: a win-win combination to easily measure prognostic
  biomarkers in {DLBCL} patients.
\newblock {\em Journal of Nuclear Medicine}, page jnumed.121.263501, June 2022.

\end{thebibliography}


\end{document}